%% file: main.tex
\documentclass{article}

\usepackage[utf8]{inputenc}
\usepackage[T1]{fontenc}
\usepackage[margin=1in]{geometry}
\usepackage[round,authoryear]{natbib}
\usepackage{amsmath}
\usepackage{amsfonts}
\usepackage{amssymb}
\usepackage{booktabs}
\usepackage{graphicx}
\usepackage{float}
\usepackage{hyperref}
\usepackage{microtype}
\usepackage{nicefrac}
\usepackage{url}
\usepackage{tikz}
\usetikzlibrary{arrows.meta,backgrounds,calc,decorations.pathreplacing,fit,
  positioning,shapes.geometric}

\DeclareRobustCommand{\system}{\textsc{ProofDiff}}

\title{\system{}: Auditable Argument-Level Comparison of AI-Generated Mathematical Proofs}

\author{%
  \textbf{Matan Tsafrir}\textsuperscript{1} \quad
  \textbf{Esdras Medeiros}\textsuperscript{2} \quad
  \textbf{Jorge Poco}\textsuperscript{3} \\
  \textbf{Kevin Gomez}\textsuperscript{1} \quad
  \textbf{Eduardo V. Teixeira}\textsuperscript{4} \quad
  \textbf{Claudio Silva}\textsuperscript{1} \\[3pt]
  \textsuperscript{1}New York University \quad
  \textsuperscript{2}Universidade Federal do Cear\'a \\
  \textsuperscript{3}Funda\c{c}\~ao Getulio Vargas \quad
  \textsuperscript{4}Oklahoma State University \\[3pt]
  \texttt{mt6037@nyu.edu} \quad
  \texttt{esdras@mat.ufc.br} \quad
  \texttt{jorge.poco@fgv.br} \\
  \texttt{kg4267@nyu.edu} \quad
  \texttt{eduardo.teixeira@okstate.edu} \quad
  \texttt{csilva@nyu.edu}
}
\date{}

\begin{document}

\maketitle

\begin{abstract}
Mathematical understanding depends on the route from hypotheses to conclusion:
how the parts of a proof function and which dependencies carry the argument
forward. This structure can persist when wording or notation changes, while a
small alteration to a hypothesis or inference can produce a different argument
beneath similar prose. Formal verification can establish that a rewrite proves
the same theorem, but not whether it follows the same mathematical route.
Comparing informal proofs, especially those produced or revised by generative
models, therefore requires reconstructing which claims, dependencies, and
reasoning steps are preserved, reorganized, or replaced.

We present \system{}, a system for comparing informal mathematical documents.
It is designed to support human review, not to certify mathematical
correctness. Given two documents, \system{} independently constructs a
source-grounded argument map for each and proposes correspondences between the
resulting maps. Each map captures the document's claims and reasoning steps and
records the typed relations that connect them. An integrated human-review layer
allows reviewers to revise the maps and correspondences while preserving an
auditable history. Comparing the two argument maps provides a
basis for inspecting whether the documents follow the same mathematical route
and where their claims or dependencies differ. We demonstrate \system{} on
four illustrative pairs of AI-generated proofs, ranging from a detailed proof
and its concise revision to proofs using different mathematical strategies.
\end{abstract}

\section{Introduction}

Thurston locates mathematical progress in the growth of human understanding
\citep{thurston1994proof}. Seen from this perspective, the label \emph{true}
records a destination, while the proof carries the mathematical route. Much of
the substance communicated by a paper lies in that route, in the logical
movement from hypotheses to conclusion. Each component acquires significance
through the role it plays and the dependencies that bind the argument together.

A proof may remain recognizably the same after every sentence has been rewritten
or its notation replaced. Conversely, a small change to a hypothesis or
dependency may alter the argument while leaving most of the prose intact. An
argument-level comparison should recognize corresponding claims and reasoning
steps even when they are expressed differently or divided at different levels
of granularity. It should provide evidence for distinguishing shared reasoning
from components that merely play analogous argumentative roles. It should also
expose changes in claims, inferences, or dependencies that similar wording may
conceal.

Standard text-comparison tools operate at a different level. They can identify
shared wording and changed passages, but they are not designed to recover the
roles or dependency relationships that give those passages their place in a
mathematical argument.

As generative AI becomes increasingly involved in mathematical research, its
mathematical output must remain intelligible and open to human scrutiny. One way
to support such understanding is to inspect how the mathematical argument
changes across successive versions of a paper. Even when the arguments are
formalized, the need to understand their relationship remains: a proof assistant
can verify each derivation against the same stated theorem, but two successful
checks do not reveal whether the derivations follow the same mathematical route.

\paragraph{Contribution.}
Our principal technical contribution is the design and implementation of
\system{}, an integrated system for comparing informal mathematical documents.
Given two documents, \system{} independently constructs a source-grounded
argument map for each and proposes correspondences between the resulting maps.
Each map captures the document's claims and reasoning steps through nodes
connected by typed relations. Comparing the two argument maps provides a basis
for inspecting whether the documents follow the same mathematical route and
where their claims or dependencies differ. Proposed correspondences
may connect individual nodes or groups of nodes across the maps, accommodating
differences in granularity. Each proposal records its provenance, a textual
proposal note, and a heuristic confidence score.

\system{} treats its automatic output as a starting point for human review
rather than as a final determination. Its integrated human-review layer allows
reviewers to revise nodes, dependencies, and alignments. The system does not
certify mathematical correctness or determine novelty.

We demonstrate \system{} on four illustrative pairs of AI-generated proofs,
ranging from a detailed proof and its concise revision to reformulations of a
shared blueprint and proofs with different mathematical machinery. These cases
demonstrate feasibility rather than general performance.

\section{Related work}
\label{sec:related}

Mathematical information retrieval focuses on finding relevant answers and
formulae \citep{mansouri2022arqmath}, while statement-to-proof matching asks
which proof belongs to a given mathematical statement \citep{li2023bert}.
TheoremGraph constructs statement-level dependency graphs at corpus scale and
connects formal and informal mathematical resources
\citep{kurgan2026theoremgraph}. These tasks concern retrieving mathematical
content or representing dependencies across collections. \system{} instead
compares the internal argument structures of two documents.

We claim no originality for typed dependency graphs over mathematics.
Unlike TheoremGraph, which stores proofs with their statements, our
representation treats proofs as nodes in their own right. This allows a shared
statement and differing proofs to be represented separately in pairwise
comparison. Relations that do not justify an inference are retained as
\texttt{context} rather than discarded, since explanatory prose can still carry
argumentative content. TheoremGraph's informal graph labels edges by the
extractor that proposed them; our relation types describe mathematical roles,
with provenance stored separately.

Argument mining identifies argumentative components and their relations within
scientific papers \citep{lauscher2018sciarg} and across documents
\citep{ruosch2025interdocument}. \system{} applies this structural perspective
to mathematical statements, proof steps, and their typed relations.
Evaluations of model-generated mathematical solutions likewise emphasize
complete reasoning rather than final answers alone \citep{petrov2025proof}.
\system{} supports such inspection through source-grounded argument maps and
correspondences but does not evaluate proof correctness.

Formal methods address related comparison and revision tasks. Regression
verification seeks to establish equivalence between successive program versions
\citep{godlin2009regression}. Proof-repair methods transport Coq proofs across
type equivalences \citep{ringer2021proofrepair}, while Baldur uses verifier
error messages to repair generated Isabelle proofs \citep{first2023baldur}.
The Lean Refactor Arena evaluates rewritten Lean proofs that retain the same
theorem statement \citep{leanrefactorarena2026}. In these settings, machine
checkers can establish correctness or equivalence, but their verdicts do not reveal
whether two proofs follow the same mathematical route. \system{} instead
constructs source-grounded argument maps and candidate correspondences for human
inspection of informal documents; it neither formalizes nor certifies the
proofs.

Document-processing systems recover structured content from scholarly PDFs
\citep{grobid,blecher2024nougat}; specialized multimodal methods extract
theorem-like statements and proofs from mathematical papers
\citep{mishra2024multimodal}. \system{} incorporates this upstream task into its
pipeline for constructing and comparing argument maps.

\section{Argument-level comparison: definitions and objective}
\label{sec:task}

We define the objects used to represent each document, its internal argument
structure, and the correspondences proposed or curated across documents.

\paragraph{Document.}
Let $D_o$ and $D_f$ be two mathematical documents, labelled
\emph{original} and \emph{final} for orientation. These labels do not assert
chronology or influence. Each document may be supplied as TeX, PDF, or both;
Section~\ref{sec:ingestion} defines the source-selection and grounding policy.

\paragraph{Node.}
For $d\in\{o,f\}$, let $V_d$ be the set of nodes representing $D_d$. A node
$v\in V_d$ represents one argument unit and records its mathematical content
and its role in the document. Its fields include a semantic type, an informative
name, a claim or description, assumptions, and techniques. Node types include
theorems, lemmas, definitions, proof steps, constructions, and conceptual
explanations. Formal statements and their proofs are represented as separate
nodes.

\paragraph{Source span.}
A source span is a record that links a node to the evidence in the input document
from which it was constructed. Every node refers to one or more source spans.
The span is the evidential anchor for the node; a model-written summary is not a
replacement for that evidence.

\paragraph{Typed relation.}
Let
\[
    \mathcal{K}=\{\texttt{proves},\texttt{uses\_result},
    \texttt{unfolds\_definition},\texttt{context}\}.
\]
A typed relation in $D_d$ is a triple $(u,k,v)\in
V_d\times\mathcal{K}\times V_d$, directed from the source node $u$ to the
target node $v$. We write
\[
    R_d\subseteq V_d\times\mathcal{K}\times V_d
\]
for the set of all such relations in $D_d$. A \texttt{proves} relation links a
proof node to the statement it proves. A \texttt{uses\_result} relation means
that $v$ uses a result or proof-specific construction supplied by $u$ to justify
an inference. An \texttt{unfolds\_definition} relation means that $v$ invokes
or checks defining conditions supplied by $u$. A \texttt{context} relation
records vocabulary, notation, or ambient structure that does not itself justify
an inference. Each non-structural relation carries evidence and provenance.

\paragraph{Dependency.}
A logical dependency is a relation $(u,k,v)\in R_d$ whose type satisfies
\[
    k\in\{\texttt{uses\_result},\texttt{unfolds\_definition}\}.
\]
We write it as $u\to v$, meaning that $v$ depends on mathematical content supplied by $u$.
Neither \texttt{context} nor \texttt{proves} is a logical dependency.
Dependencies are direct: if $A\to B$ and $B\to C$, then $A\to C$ is included
only when $C$ also uses $A$ directly. A forward dependency is one whose source
occurs later than its target in reading order.

\paragraph{Argument map.}
The argument map of $D_d$ is the typed directed graph
\[
    G_d=(V_d,R_d).
\]
Its logical dependency graph is the projection of $R_d$ onto
\texttt{uses\_result} and \texttt{unfolds\_definition}. This projection is a
view of the same relation store, not a second dependency representation. The
comparison is defined between the two argument maps $G_o$ and $G_f$.

\paragraph{Candidate alignment.}
A candidate alignment is a proposed cross-document correspondence
\[
    a=(O,F,s,r,c,\pi),
    \qquad
    \varnothing\ne O\subseteq V_o,\quad
    \varnothing\ne F\subseteq V_f,
\]
where $O$ and $F$ are the proposed corresponding node groups, $r$ is a
free-text proposal note recording the basis or source grounding of the
correspondence,
$c\in[0,1]$ is a proposer-specific heuristic confidence score, and $\pi$
records proposal provenance. The confidence score is not interpreted as a
calibrated probability or an acceptance decision.
The shape $s$ is \texttt{one\_to\_one}, \texttt{split}, \texttt{merge}, or
\texttt{many\_to\_many} according to whether the endpoint cardinalities are
$(1,1)$, $(1,n)$, $(n,1)$, or $(n,m)$ with $n,m>1$, respectively. Grouped
endpoints represent a possible difference in argument granularity. Every
candidate remains traceable to source evidence through its endpoint nodes, and
a node may occur in several candidates when they represent different possible
correspondences.

\paragraph{Unresolved node.}
At any stage of review, a node that is not covered by an active automatic
candidate or human-curated alignment remains \texttt{unresolved}. An unresolved
node becomes \texttt{unique} only through an explicit human-review decision.
Failure to retrieve or retain a correspondence is therefore not evidence of
novelty.

\paragraph{Comparison objective.}
Given $D_o$ and $D_f$, \system{} constructs source-grounded argument maps $G_o$
and $G_f$, proposes a traceable candidate set $A$, and allows reviewers to
revise nodes, dependencies, and alignments. These artifacts support inspection
of where the arguments correspond or remain unresolved. Neither an alignment
nor its absence is a verdict about correctness, novelty, or influence.

\section{\system{}'s automatic pipeline}
\label{sec:pipeline}

\system{} takes two mathematical documents as input and independently
constructs a source-grounded argument map for each. It then proposes candidate
alignments between the maps.

\subsection{TeX-first ingestion and source grounding}
\label{sec:ingestion}

\system{} accepts each document as TeX, PDF, or both. When TeX is available,
the system uses it as the source for later analysis. A deterministic TeX parser
reads the main source file and its included files and divides the source into
structural blocks. Each block retains the raw TeX, cleaned text, source file,
and line numbers. These blocks are the source evidence from which nodes are
constructed.

When TeX compiles, SyncTeX maps each source block to its rendered page region.
For PDF-only input, a layout-aware extractor creates text blocks and retains
their page regions.

\subsection{Node construction and enrichment}

Node construction starts from the source blocks produced during
ingestion. First, deterministic rules use the document's section structure to
identify named results, proofs, and explicit proof steps. Each becomes a
preliminary node, as does substantive prose outside these units.

Next, \system{} makes one structured model call for each top-level section. The
call receives the preliminary nodes and their corresponding source text. The
model assigns informative names and describes each node's mathematical content,
role, assumptions, and techniques. It may also correct a node type or divide a
coarse proof into contiguous steps, while every resulting node remains linked
to its original source spans.

The two documents are processed independently: neither document's segmentation
nor node descriptions inform the other's map. Appendix~\ref{app:prompts} gives
the prompts and schemas.

\subsection{Typed relation discovery}

After the nodes and their descriptions have been finalized, relation discovery
determines how the nodes connect to form an argument map.

For each document section identified during the preceding deterministic pass, a
first model call receives the source text of the nodes in that section together
with an index of every node in the document. The model proposes plausible direct
relations, including relations that cross section boundaries. The pipeline
combines these with candidates from explicit TeX references and relations
already in the map, validates endpoints, and removes duplicates.

When candidates are present, a second model call classifies each as
\texttt{uses\_result}, \texttt{unfolds\_definition},
\texttt{context}, or \texttt{irrelevant}. The pipeline discards candidates
classified as \texttt{irrelevant} and stores an explanation and provenance for
each retained relation. Proof-to-statement relations are not classified by the
model. Instead, the pipeline uses the document outline to link each proof node
to the statement it proves and records the relation as \texttt{proves}.
Appendix~\ref{app:prompts} describes the prompts and additional implementation
details.

\subsection{Cross-document candidate alignment}

After both argument maps have been constructed, the automatic alignment stage
generates candidate correspondences from three sources: deterministic anchors,
a model call, and optional imports.

Deterministic anchors use result type and title similarity, and separately
match associated proofs.

A single structured model call receives the semantic descriptions of every
node in both maps. The model may propose one-to-one or grouped correspondences
and returns a rationale and confidence score for each proposal. The model does
not receive the typed relations or graph topology; its proposals are based on
the node descriptions. An optional import mechanism can also convert externally
supplied pairs of quoted passages into candidates by mapping those passages to
source blocks and then to nodes.

Every candidate records its proposal source and a textual proposal note. For
model-generated candidates, this note gives the rationale for the proposed
correspondence, allowing a reviewer to inspect whether it reflects shared
reasoning or only a similar argumentative role; \system{} does not classify
that distinction automatically.

Typed relations remain available as additional evidence for visualization
and human inspection, but the current candidate-generation and scoring
mechanisms do not use them. Incorporating structural agreement into candidate
generation or scoring remains future work.

Confidence is assigned by the mechanism that produces each candidate.
Deterministic anchors and imported comparisons use fixed formulas based on
title similarity or passage coverage, while the model returns its own heuristic
score. These proposer-specific values are uncalibrated and should not be
interpreted as probabilities or acceptance decisions. Confidence filters
affect display only.

The automatic alignment stage ends with this candidate set and does not
adjudicate its proposals. Section~\ref{sec:master-documents} describes the
subsequent human-review layer. Appendix~\ref{app:prompts} describes the model
prompt, output schema, and confidence calculations.

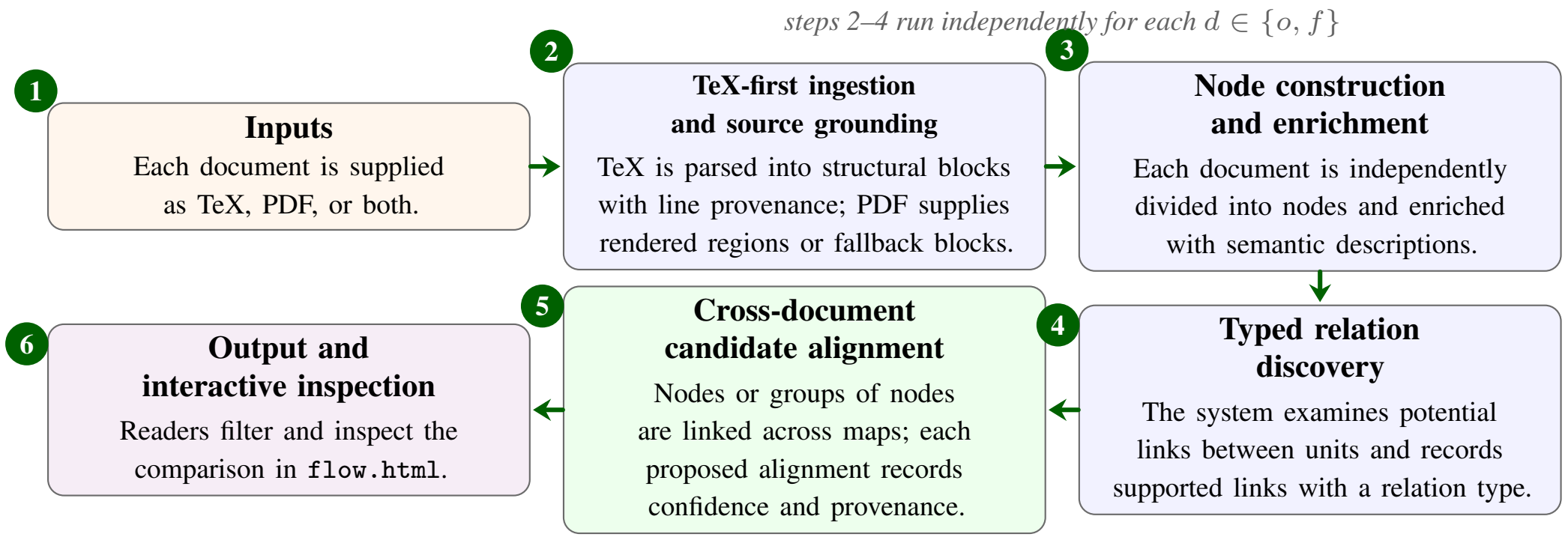
\begin{figure}[H]
  \centering
  {\fontfamily{ptm}\selectfont
  \input{figures/pipeline-overview.tex}}
  \caption{The automatic \system{} pipeline. Each document is processed
  independently before cross-document candidate alignment; the resulting maps
  and proposals are presented in \texttt{flow.html}.}
  \label{fig:pipeline-overview}
\end{figure}

Figure~\ref{fig:pipeline-overview} summarizes the automatic pipeline described
above.

\section{Output, interactive inspection, and human curation}
\label{sec:output}

\subsection{Output and interactive inspection}

\begin{figure}[H]
  \centering
  \includegraphics[width=0.95\linewidth]{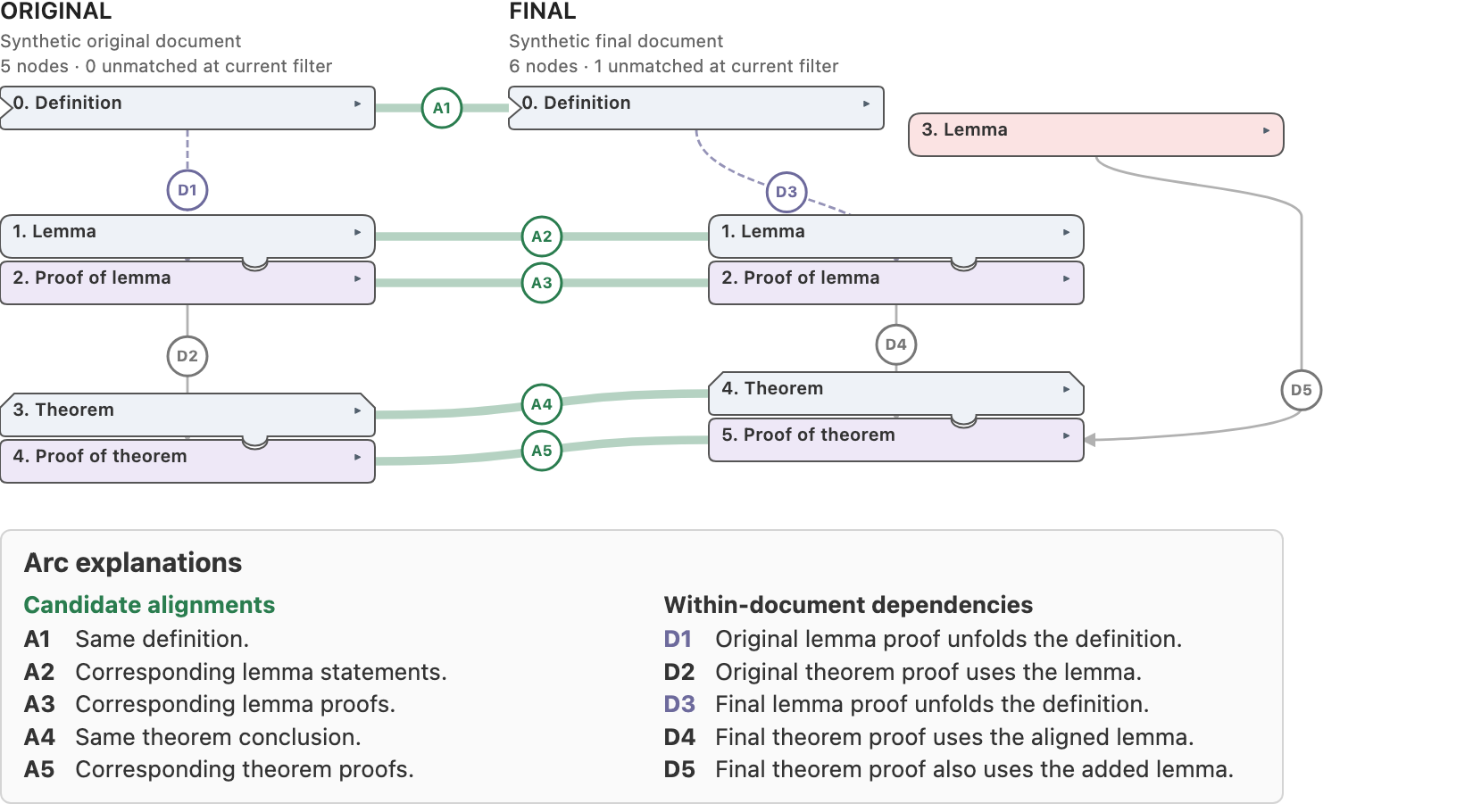}
  \caption{An annotated synthetic comparison rendered by the production
  \texttt{flow.html} interface. A1--A5 mark candidate alignments and D1--D5
  mark within-document dependencies; the key explains each arc. The additional
  final lemma has no candidate correspondence and remains unresolved.}
  \label{fig:argument-map-example}
\end{figure}

The automatic pipeline produces two source-grounded argument maps $G_o$ and
$G_f$ and a set $A$ of candidate alignments between them. The browser-based
\texttt{flow.html} interface renders the maps as a side-by-side interactive
diagram, with nodes for argument units, edges for within-document relations,
and links between maps for candidate alignments. Users can switch between
reading-order and dependency-graph layouts, expand proofs, filter alignments,
and isolate unmatched material.
Figure~\ref{fig:argument-map-example} illustrates the main elements of the
comparison view.

Selecting a node opens a side panel containing its mathematical description,
source evidence, typed dependencies, and cross-document alignments, together
with their explanations, provenance, confidence scores, and review status.
Nodes without a visible automatic or human-curated alignment are displayed as
unmatched; this does not classify them as unique or novel.

Appendix~\ref{app:interface} describes the interface controls and inspection
workflow.

\subsection{Human curation and editable master documents}
\label{sec:master-documents}

\system{}'s human-review layer allows reviewers to add, modify, reject, or
restore dependencies and alignments while preserving the automatic output as
underlying evidence.

For a TeX-backed comparison, the curated state is stored in three synchronized
editable master documents: an annotated TeX master for each input and a
pair-level alignment master. Comment-only annotations in the TeX masters record
node boundaries, descriptions, and dependencies without altering the paper text
or rendering. Reviewers can edit node information through these masters and
curate dependencies and alignments through either the masters or the interface.

For such TeX-backed comparisons, each approved edit updates the active
comparison and creates an immutable pair-level revision, while the initial
automatic output remains available as a restorable baseline.
Appendix~\ref{app:interface} describes validation, synchronization, and revision
history in detail.

\section{Comparisons on generated proof pairs}
\label{sec:examples}

\subsection{Example set and construction}

We illustrate \system{} on four pairs of AI-generated informal mathematical
proofs, all supplied as TeX. The pairs span preservation under reformulation,
compression, different proof strategies, and broader functional
correspondence.

The eight proofs were generated through consumer-facing interfaces using
Claude, Gemini 3.1 Pro, and GPT-5.6 Sol; all model-backed stages within
\system{} used GPT-5.5. Appendix~\ref{app:toy-inputs} describes the construction
of the proof pairs and the model settings.

All reported comparisons use automatic output from TeX inputs, with no imported
candidates or saved human edits. The figures use the illustrative threshold
$c\geq 0.75$, which retains 12 of 21 candidates
(Table~\ref{tab:summary}). Because the confidence scores are uncalibrated, this
threshold is a display filter rather than an acceptance criterion. We report
all 21 candidates. The authors inspected all 21 candidates without independent
evaluation; the examples show feasibility, not representative performance.

\begin{table}[H]
  \centering
  \small
  \begin{tabular}{lccc}
    \toprule
    Pair & Nodes O/F & Candidates all/filtered & Unmatched O/F \\
    \midrule
    Infinitude of primes           & 5/4  & 4/4 & 1/0 \\
    Cauchy--Schwarz                & 5/5  & 6/2 & 3/3 \\
    Uncountability of $\mathbb{R}$ & 7/10 & 6/2 & 5/8 \\
    Mean value theorem             & 11/4 & 5/4 & 2/0 \\
    \bottomrule
  \end{tabular}
  \caption{Automatic output on the four pairs at $c\geq 0.75$. O/F denotes
  original/final. Candidate counts are proposals, not accepted alignments.}
  \label{tab:summary}
\end{table}

\subsection{Infinitude of primes: preservation under reformulation}

For this preservation case, Claude Opus 5 generated the first proof and then
extracted its argument blueprint in a follow-up prompt. Gemini 3.1 Pro generated
the second proof under that blueprint, preserving the mathematical route while
changing its expression.

\begin{figure}[H]
  \centering
  \includegraphics[width=0.95\linewidth]{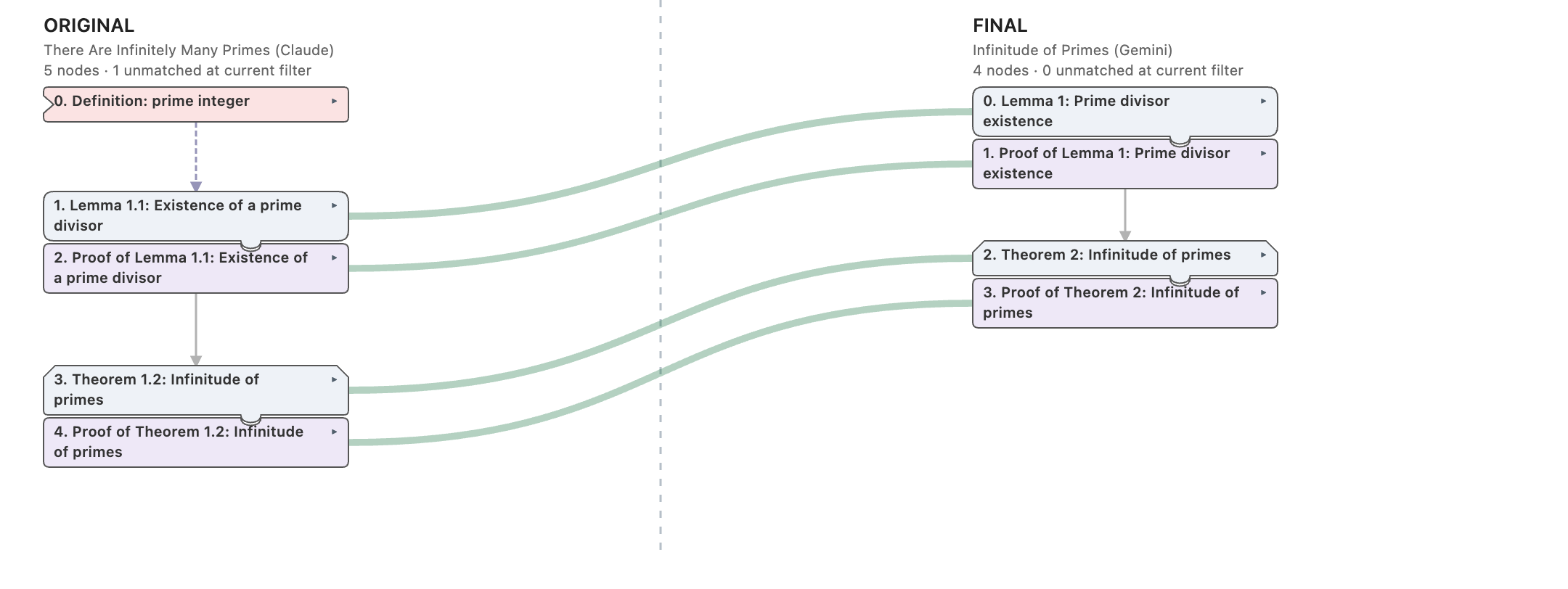}
  \caption{Automatic \system{} flow view for the infinitude-of-primes pair with
  proof nodes expanded and confidence filter $c\geq 0.75$. The original's
  explicit prime definition is the only unresolved node.}
  \label{fig:infinitude-primes-flow}
\end{figure}

At the confidence threshold $c\geq 0.75$, all four candidate alignments remain,
linking the corresponding auxiliary lemmas, theorems, and their proofs; each was
proposed by both the deterministic anchor and model retrieval. The original's
explicit definition of a prime has no separate counterpart and remains unresolved
(Figure~\ref{fig:infinitude-primes-flow}).

\subsection{Cauchy--Schwarz: shared statement, different strategy}
\label{sec:cauchy}

Claude Opus 5 generated two proofs of Cauchy--Schwarz using different methods:
a discriminant argument and normalization to unit vectors. The pair tests
whether \system{} aligns their shared setting and conclusion while keeping
method-specific steps distinct.

\begin{figure}[H]
  \centering
  \includegraphics[width=0.95\linewidth]{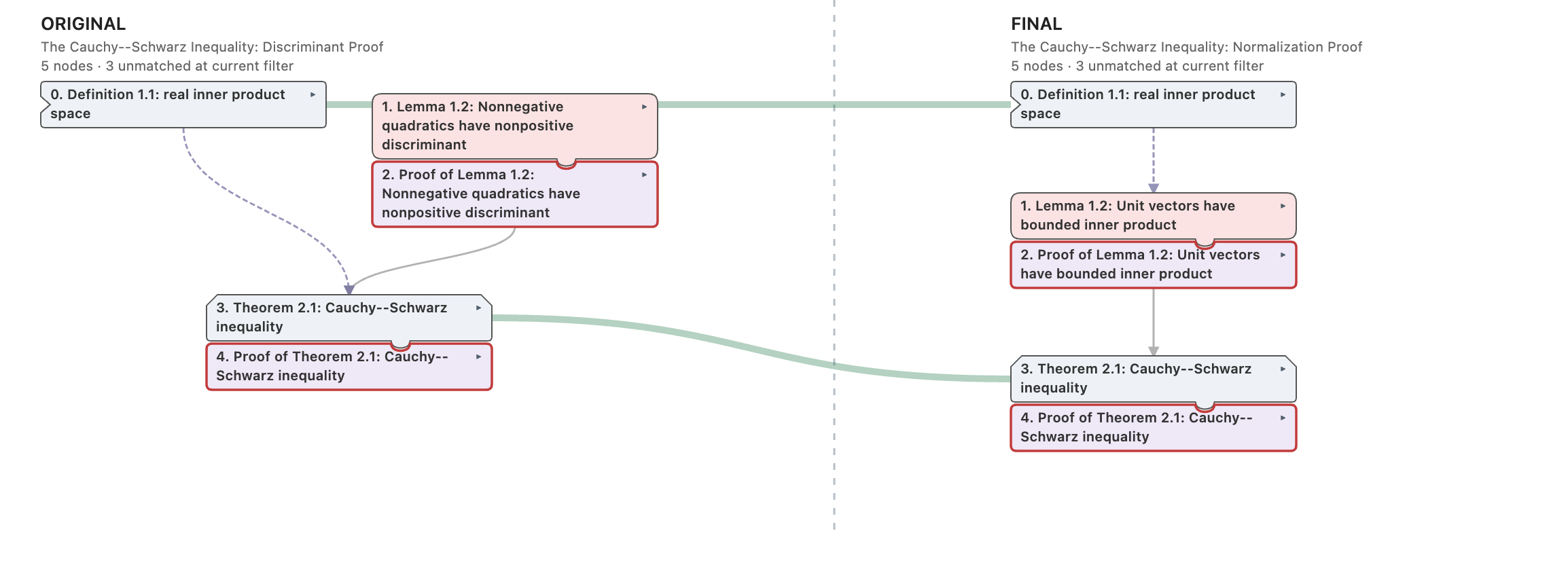}
  \caption{Automatic flow view with proof nodes expanded at $c\geq 0.75$;
  only the shared definition and theorem statement remain visible
  correspondences, while the method-specific lemmas and proofs remain
  unresolved.}
  \label{fig:cauchy-schwarz-flow}
\end{figure}

At $c\geq 0.75$, two of six candidates remain, aligning the common definition
and theorem statement. The discriminant and normalization arguments do not
appear as a strong proof correspondence, and their method-specific
lemma--proof pairs stay unresolved
(Figure~\ref{fig:cauchy-schwarz-flow}).

Below the threshold, type-uniqueness anchors pair the method-specific lemmas and
proofs at $c=0.60$ and $0.15$. Model proposals record partial common ground:
the original theorem proof and the final unit-vector proof both expand
nonnegative squared norms ($c=0.52$), while the two theorem proofs share only
limited outer structure ($c=0.34$). Their central strategies remain different
(Appendix~\ref{app:candidates}).

\subsection{Uncountability of the real numbers: shared conclusion, different
machinery}
\label{sec:uncountability}

Claude Opus 5 generated two proofs of the uncountability of $\mathbb{R}$: a
diagonal argument and a measure-theoretic argument. The pair tests a comparison
in which the conclusion is shared but the intermediate machinery and argument
structure differ.

\begin{figure}[H]
  \centering
  \includegraphics[width=0.95\linewidth]{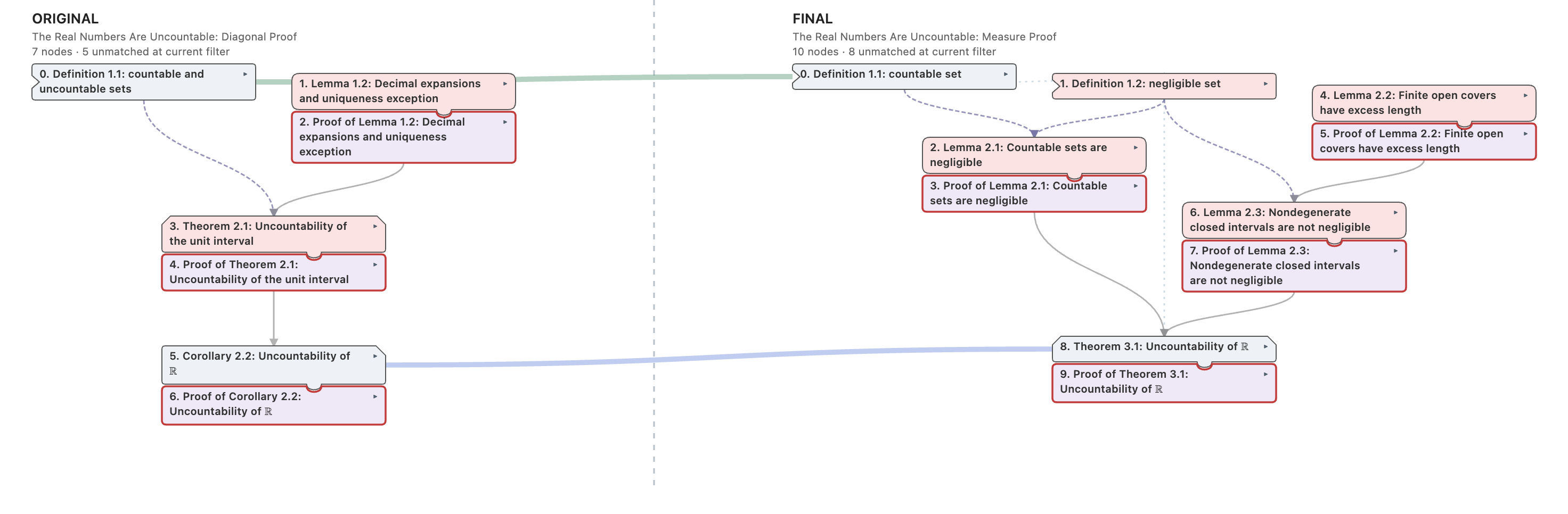}
  \caption{Automatic flow view with proof nodes expanded at $c\geq 0.75$.
  The common definition and conclusion remain visible correspondences, while
  the diagonal and measure-theoretic proof machinery remains unresolved.}
  \label{fig:uncountability-flow}
\end{figure}

At $c\geq 0.75$, two of six candidates remain: the countability definitions
and the common conclusion. \system{} matches the original corollary with the
final theorem despite their different formal labels. The diagonal and
measure-theoretic proofs of that conclusion remain below the threshold, as do
the other method-specific nodes
(Figure~\ref{fig:uncountability-flow}).

Below the threshold, a grouped proposal at $c=0.52$ connects the original
unit-interval theorem to the two measure-theoretic lemmas that jointly provide
an interval-level obstruction. A separate proposal at $c=0.58$ records the
shared contradiction-and-restriction structure of the concluding proofs while
keeping their diagonal and measure-theoretic machinery distinct. Type-uniqueness
anchors also produce weaker cross-level candidates at $c=0.60$ and $0.15$
(Appendix~\ref{app:candidates}).

\subsection{Mean value theorem: compression across levels of detail}
\label{sec:mvt}

Claude generated a detailed, undergraduate-exam-level proof of Lagrange's mean
value theorem. The proof was then given to GPT-5.6 Sol with a request to make it
concise while preserving completeness and correctness, creating a test of
alignment across levels of granularity.

\begin{figure}[H]
  \centering
  \includegraphics[width=0.95\linewidth]{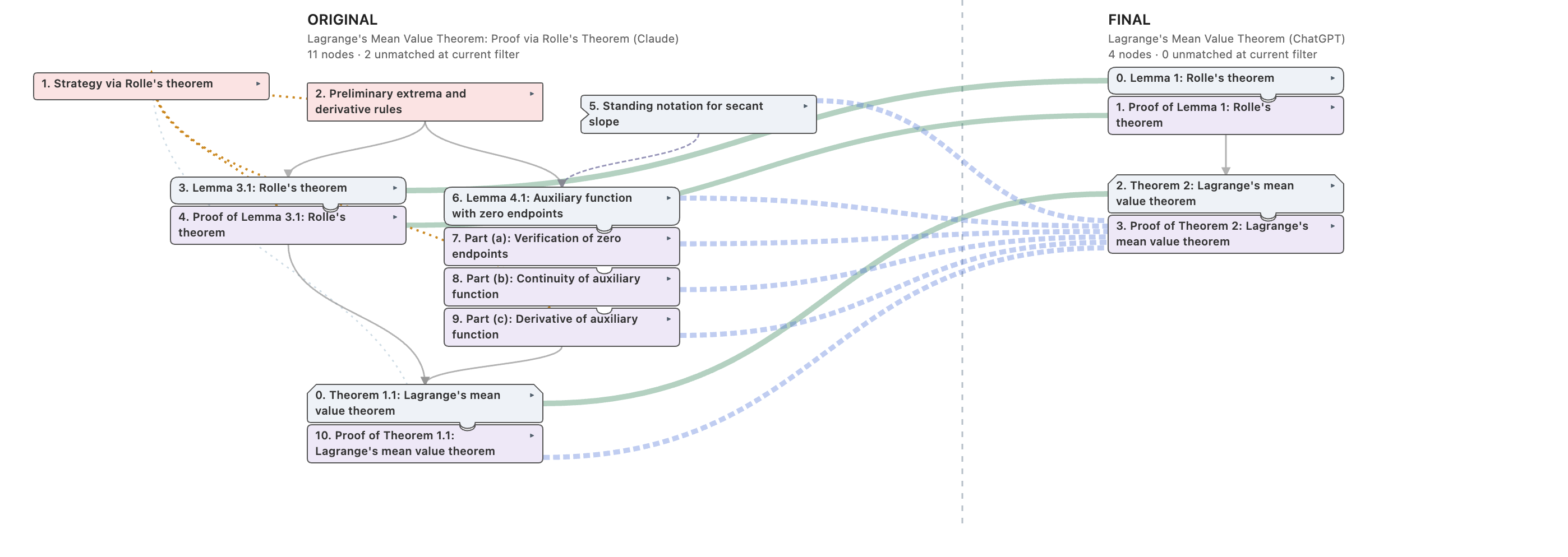}
  \caption{Automatic flow view with proof nodes expanded at $c\geq 0.75$.
  A grouped proposal connects the secant-slope notation, auxiliary-function
  statement and its three proof parts, and final Rolle application to the
  concise proof; the Rolle statement and proof also align.}
  \label{fig:mvt-flow}
\end{figure}

At $c\geq 0.75$, four of five candidates remain. The mean value and Rolle
statements align, as do the Rolle proofs. A grouped proposal connects the
original secant-slope notation, auxiliary-function statement, its three
labelled proof parts, and the final application of Rolle's theorem with
ChatGPT's single concise proof node. This many-to-one correspondence represents
the compression of material that the original presents as several separate
argument units (Figure~\ref{fig:mvt-flow}).

The remaining candidate is a low-confidence direct proof pairing at $c=0.25$
and stays below the display threshold.

Together, these cases show that \system{} can expose shared argument structure
across reformulations and granularity changes while leaving method-specific
structure unresolved. They validate neither the threshold nor recall.

\section{Limitations and outlook}
\label{sec:limitations}

This early implementation establishes feasibility rather than measuring
performance. Broader claims require independent expert annotation, more varied
documents, recall evaluation, and confidence calibration. All evaluated
documents are short and AI-generated; human-written proofs, revisions of real
mathematical texts, and longer documents remain to be studied.

Model-generated descriptions, dependencies, and candidate alignments may omit
structure, propose unsupported relationships, or vary across runs. Source
grounding, provenance, and human curation make these outputs inspectable but do
not eliminate error or the need for human judgment. The system also does not
yet classify aligned material as preserved, reorganized, or differently reasoned.

\system{} does not establish proof correctness, novelty, authorship, or causal
influence. Formal verification remains complementary; broader conclusions
require further validation.

\section{Conclusion}

\system{} compares mathematical reasoning through independently constructed
argument maps and source-grounded candidate alignments. Its human-review layer
allows reviewers to revise nodes, dependencies, and alignments while preserving
provenance and revision history. Across our illustrative examples, the
automatic output exposes shared structure while leaving method-specific
material unresolved. This proof of concept is an early step toward making
mathematical change legible across human- and model-generated proofs.

\section*{Publication Note}

An earlier version of this work was accepted by the NeurIPS 2026 Workshop on
AI for Verifiable Coding.

\section*{Artifact Availability}

ProofDiff v0.1.1 is available from the public release at
\url{https://github.com/VIDA-NYU/proofdiff-public/releases/tag/v0.1.1}.
The exact software snapshot, example inputs, archived outputs, sanitized run
metadata, interactive views, and paper figures are preserved at
\url{https://doi.org/10.5281/zenodo.23930759}.
Model-backed stages are nondeterministic, so later runs may differ even with
the same inputs and settings.

\section*{Acknowledgements and Disclosure of Funding}

This work was funded by the DARPA expMath program (Agreement
No.~HR0011262E029). Silva was also partially supported by NSF grants
OAC-2411221 and DRL-2604563. Poco was also partially supported by the
Brazilian National Council for Scientific and Technological Development
(CNPq), grant \#313454/2026-4, and the Getulio Vargas Foundation (FGV). Any
opinions, findings, conclusions, or recommendations expressed in this material
are those of the authors and do not necessarily reflect the views of DARPA,
NSF, or the authors' affiliated institutions. The authors declare no competing
interests.

\bibliographystyle{plainnat}
\bibliography{references}

\appendix

\input{appendix.tex}

\section{Toy proof inputs}
\label{app:toy-inputs}

The archived artifact at \url{https://doi.org/10.5281/zenodo.23930759}
contains the eight AI-generated informal proofs used in
Section~\ref{sec:examples}, compiled as PDFs from the exact TeX inputs given to
\system{}. The labels \emph{original} and \emph{final} indicate only the
orientation of each comparison. The Claude proofs in the first three pairs were
generated with Claude Opus 5 through Claude.ai at the \emph{High} effort
setting; Gemini 3.1 Pro was accessed through Gemini Apps with \emph{Extended}
thinking enabled. The mean-value-theorem revision was generated with GPT-5.6
Sol through ChatGPT.

\begin{table}[H]
  \centering
  \small
  \begin{tabular}{p{0.24\linewidth}p{0.32\linewidth}p{0.32\linewidth}}
    \toprule
    Pair & Original PDF & Final PDF \\
    \midrule
    Infinitude of primes
      & Claude Opus 5 proof generated first
      & Gemini 3.1 Pro proof generated from its blueprint \\
    Cauchy--Schwarz
      & Discriminant proof
      & Normalization proof \\
    Uncountability of $\mathbb{R}$
      & Diagonal proof
      & Measure-theoretic proof \\
    Mean value theorem
      & Detailed Claude proof
      & GPT-5.6 Sol revision prompted to remain complete while becoming concise \\
    \bottomrule
  \end{tabular}
  \caption{The proof inputs distributed under
  \texttt{toy\_proof\_inputs/<pair>/original.pdf} and
  \texttt{final.pdf} in the archived artifact.}
  \label{tab:toy-inputs}
\end{table}

\section{Complete candidate proposals}
\label{app:candidates}

The following tables list all 21 automatic candidate alignments in the four
proof-pair comparisons. The 12 candidates with $c\geq 0.75$ enter the filtered
views in Section~\ref{sec:examples}. Proposal provenance is abbreviated as A
(deterministic anchor) and M (model retrieval). The final column gives the
authors' short interpretation used in the case discussion; it is not an
independently adjudicated ground-truth label.

\begin{table}[H]
  \centering
  \scriptsize
  \setlength{\tabcolsep}{3pt}
  \begin{tabular}{cp{0.25\linewidth}p{0.25\linewidth}ccc}
    \toprule
    ID & Original endpoint & Final endpoint & Source & Conf. & Interpretation \\
    \midrule
    1 & Existence of a prime divisor & Prime divisor existence
      & A+M & .98 & supported \\
    2 & Minimal divisor is prime & Minimal divisor is prime
      & A+M & .93 & supported \\
    3 & Infinitude theorem & Infinitude theorem & A+M & .99 & supported \\
    4 & Euclid contradiction from a finite list
      & Euclid contradiction from product plus one
      & A+M & .95 & supported \\
    \bottomrule
  \end{tabular}
  \caption{All candidates for the infinitude-of-primes pair.}
  \label{tab:candidates-primes}
\end{table}

\begin{table}[H]
  \centering
  \scriptsize
  \setlength{\tabcolsep}{3pt}
  \begin{tabular}{cp{0.25\linewidth}p{0.25\linewidth}ccc}
    \toprule
    ID & Original endpoint & Final endpoint & Source & Conf. & Interpretation \\
    \midrule
    1 & Inner-product-space definition & Inner-product-space definition
      & A+M & 1.00 & supported \\
    2 & Nonnegative-quadratic lemma & Unit-inner-product lemma
      & A & .60 & unsupported \\
    3 & Completing-square proof & Proof of the unit-vector bound
      & A & .15 & unsupported parent hint \\
    4 & Cauchy--Schwarz theorem & Cauchy--Schwarz theorem
      & A+M & 1.00 & supported \\
    5 & Quadratic proof of Cauchy--Schwarz
      & Proof of the unit-vector bound
      & M & .52 & partial shared norm expansion \\
    6 & Quadratic proof of Cauchy--Schwarz
      & Proof by normalization to unit vectors
      & A+M & .34 & weak outer structure; different method \\
    \bottomrule
  \end{tabular}
  \caption{All candidates for the Cauchy--Schwarz pair.}
  \label{tab:candidates-cs}
\end{table}

\begin{table}[H]
  \centering
  \scriptsize
  \setlength{\tabcolsep}{3pt}
  \begin{tabular}{cp{0.25\linewidth}p{0.25\linewidth}ccc}
    \toprule
    ID & Original endpoint & Final endpoint & Source & Conf. & Interpretation \\
    \midrule
    1 & Countable/uncountable-set definition & Countable-set definition
      & A+M & 1.00 & supported \\
    2 & Unit-interval theorem & Countable-set + nonnegligible-interval lemmas
      & M & .52 & partial shared intermediate role \\
    3 & Unit-interval theorem & Uncountability-of-$\mathbb{R}$ theorem
      & A & .60 & unsupported \\
    4 & Diagonal proof for $[0,1)$
      & Negligibility proof for $\mathbb{R}$
      & A & .15 & unsupported parent hint \\
    5 & Uncountability-of-$\mathbb{R}$ corollary
      & Uncountability-of-$\mathbb{R}$ theorem
      & M & .96 & supported \\
    6 & Restrict an enumeration to $[0,1)$
      & Contradiction using negligibility of $[0,1]$
      & M & .58 & partial shared reduction; different method \\
    \bottomrule
  \end{tabular}
\caption{All candidates for the uncountability-of-$\mathbb{R}$ pair.}
  \label{tab:candidates-uncountability}
\end{table}

\begin{table}[H]
  \centering
  \scriptsize
  \setlength{\tabcolsep}{3pt}
  \begin{tabular}{cp{0.25\linewidth}p{0.25\linewidth}ccc}
    \toprule
    ID & Original endpoint & Final endpoint & Source & Conf. & Interpretation \\
    \midrule
    1 & Mean value theorem & Mean value theorem
      & A+M & 1.00 & supported \\
    2 & Rolle's theorem & Rolle's theorem
      & A+M & 1.00 & supported \\
    3 & Extrema-and-Fermat proof of Rolle's theorem
      & Proof of Rolle's theorem
      & A+M & .91 & supported \\
    4 & Secant-slope notation + auxiliary-function statement + three proof
      parts + final Rolle application
      & Secant-line reduction to Rolle's theorem
      & M & .88 & supported grouped compression \\
    5 & Final Rolle application & Secant-line reduction to Rolle's theorem
      & A & .25 & weak parent hint; covered by group \\
    \bottomrule
  \end{tabular}
  \caption{All candidates for the detailed-to-concise mean value theorem pair.}
  \label{tab:candidates-mvt}
\end{table}

\end{document}

%% file: figures/pipeline-overview.tex
\begin{tikzpicture}[
  >={Stealth[length=4.5pt,width=5.5pt]},
  card/.style={draw=black!58, fill=white, rounded corners=4pt,
    text width=3.90cm, minimum height=1.12cm, align=center,
    font=\footnotesize, inner xsep=5pt, inner ysep=3pt},
  input card/.style={card, fill=orange!7},
  document card/.style={card, fill=blue!5},
  comparison card/.style={card, fill=green!7},
  output card/.style={card, fill=violet!7},
  step badge/.style={circle, fill=green!38!black, text=white,
    minimum size=0.38cm, inner sep=0pt, font=\scriptsize\bfseries},
  flow/.style={->, draw=green!38!black, line width=1.15pt,
    line cap=round, shorten <=0pt, shorten >=0.6pt},
  explanation/.style={font=\scriptsize, text=black!68}
]
  \node[input card] (inputs) at (-4.55,0.45)
    {{\bfseries Inputs}\\[1.5pt]
     {\scriptsize Each document is supplied\\as TeX, PDF, or both.}};
  \node[document card] (ground) at (0,0.45)
    {{\scriptsize\bfseries TeX-first ingestion\\and source grounding}\\[1.5pt]
     {\scriptsize TeX is parsed into structural blocks with line provenance;
     PDF supplies rendered regions or fallback blocks.}};
  \node[document card] (units) at (4.55,0.45)
    {{\bfseries Node construction\\and enrichment}\\[1.5pt]
     {\scriptsize Each document is independently divided into nodes and enriched with semantic descriptions.}};
  \node[document card] (maps) at (4.55,-1.70)
    {{\bfseries Typed relation\\discovery}\\[1.5pt]
     {\scriptsize The system examines potential links between units and records supported links with a relation type.}};

  \node[comparison card] (align) at (0,-1.70)
    {{\bfseries Cross-document\\candidate alignment}\\[1.5pt]
     {\scriptsize Nodes or groups of nodes are linked across maps; each proposed alignment records confidence and provenance.}};
  \node[output card] (view) at (-4.55,-1.70)
    {{\bfseries Output and\\interactive inspection}\\[1.5pt]
     {\scriptsize Readers filter and inspect the comparison in \texttt{flow.html}.}};

  \draw[flow] (inputs.east) -- (ground.west);
  \draw[flow] (ground.east) -- (units.west);
  \draw[flow] (units.south) -- (maps.north);
  \draw[flow] (maps.west) -- (align.east);
  \draw[flow] (align.west) -- (view.east);

  \node[step badge] at ([xshift=-0.10cm,yshift=0.10cm]inputs.north west) {1};
  \node[step badge] at ([xshift=-0.10cm,yshift=0.10cm]ground.north west) {2};
  \node[step badge] at ([xshift=-0.10cm,yshift=0.10cm]units.north west) {3};
  \node[step badge] at ([xshift=-0.18cm,yshift=-0.18cm]maps.north west) {4};
  \node[step badge] at ([xshift=-0.18cm,yshift=-0.18cm]align.north west) {5};
  \node[step badge] at ([xshift=-0.18cm,yshift=-0.18cm]view.north west) {6};

  \node[font=\scriptsize\itshape, text=black!62]
    at (2.28,1.72) {steps 2--4 run independently for each $d\in\{o,f\}$};
\end{tikzpicture}

%% file: appendix.tex
\section{Model prompts and their pipeline stages}
\label{app:prompts}

\system{} uses four prompt types across argument-map construction and
cross-document comparison. Each call combines a fixed system prompt defining
the task, a user prompt assembled from document artifacts, and a structured
output schema. Table~\ref{tab:prompt-stages} locates these calls using the
step numbers in Figure~\ref{fig:pipeline-overview}. The descriptions below
summarize the implemented instructions and schemas. All four prompt types were run with GPT-5.5 in the comparisons reported in
Section~\ref{sec:examples}.

\begin{table}[ht]
  \centering
  \small
  \begin{tabular}{@{}p{0.13\linewidth}p{0.29\linewidth}p{0.50\linewidth}@{}}
    \toprule
    Figure step & Prompt task & Scope and timing \\
    \midrule
    \raggedright 3: argument units & Node enrichment &
      One call per top-level section in each document, after deterministic
      segmentation. \\[3pt]
    \raggedright 4: argument maps & Relation candidate retrieval &
      One call per section in each document, after node boundaries and
      summaries have stabilized. \\[3pt]
    \raggedright 4: argument maps & Relation classification &
      One call for each section with candidate relations, after combining
      model proposals, TeX references, and existing relation evidence. \\[3pt]
    \raggedright 5: candidate alignments & Cross-document retrieval &
      One call over both constructed argument maps. \\
    \bottomrule
  \end{tabular}
  \caption{Model prompts in the pipeline. Call counts describe the normal
  execution path; validation failures can trigger correction calls.}
  \label{tab:prompt-stages}
\end{table}

\subsection{Node enrichment: step 3}

\paragraph{Input and instructions.}
The user prompt provides the semantic-source designation, draft node IDs and
types, their block ranges, and the corresponding source text tagged with
block IDs. The system prompt asks the model to annotate each node with a
short informative name, a mathematical claim, its function in the argument,
assumptions, and techniques. It instructs the model to preserve established
result names, correct a draft type only when clearly warranted, and avoid
inventing content. TeX is designated as authoritative when available; PDF
text is used when no TeX source was declared.

\paragraph{Output and constraints.}
The response contains node summaries and optional proof splits. A split must
contain two to six coherent steps whose contiguous block ranges partition
the original proof in order. Statements and explanatory prose must not be
split. Each summary records \texttt{node\_id}, \texttt{type}, \texttt{name},
\texttt{claim}, \texttt{function}, \texttt{assumptions}, and
\texttt{techniques}; each proposed step identifies its first and last source
blocks. Mathematical expressions must use delimited LaTeX with preserved
backslashes. Protected and legacy-reviewed nodes are shown for context and not
replaced. For nodes edited through the master documents, manually owned fields
and boundaries are preserved; other fields may be refreshed unless the node's
\texttt{protect} flag is enabled. Each document is enriched independently.

\subsection{Relation candidate retrieval: step 4, first pass}

The user prompt combines an index of all nodes in one document with the
source text of the target nodes in the current section. The index supplies
node IDs, types, names, and abbreviated claims, allowing candidates to refer
to nodes outside the target section. The system prompt prioritizes recall:
it asks for every plausible direct use of a result, proof-specific
construction, or defining condition, as well as contextual uses of vocabulary
or notation. Uncertain cases should be retained for classification, while
mere topical similarity and indirect, transitive relationships should be
excluded. Later source nodes are allowed as possible forward references.

The output is a list of candidate records containing \texttt{source\_id},
\texttt{target\_id}, and a short textual \texttt{cue}. This pass does not
assign relation types. The prompt excludes proof-to-statement links because
\texttt{proves} relations are supplied from the outline. Before the next
pass, the pipeline adds explicit TeX-reference candidates and existing
relation evidence, validates endpoints, and removes duplicate pairs.

\subsection{Relation classification: step 4, second pass}

The classifier receives the document index, target source text, and the
complete candidate list with its cues. Its system prompt requires exactly
one classification for every supplied pair and defines four outcomes:
\texttt{uses\_result} when mathematical content supports an inference;
\texttt{unfolds\_definition} when defining conditions are invoked or checked;
\texttt{context} for vocabulary, notation, or ambient structure; and
\texttt{irrelevant} when none applies. It asks for the strongest supported
relation, preserves implicit inferences even without an explicit citation,
and requires concrete evidence for forward references. Mentioning a defined
object alone is insufficient to establish an unfolding.

Each output record contains \texttt{source\_id}, \texttt{target\_id},
\texttt{kind}, and \texttt{evidence}. The evidence must identify the supported
inference, defining condition, or contextual role, or explain why the pair is
irrelevant. The prompt forbids invented quotations. The pipeline checks that
every candidate is classified exactly once and discards relations classified
as \texttt{irrelevant}.

Relations supported by later source passages are retained as forward
dependencies and marked in the interface. The deterministic pair validator
reports cycles among logical dependencies as errors, and the human-review layer
rejects proposed dependency edits that would introduce a cycle.

\subsection{Cross-document alignment: step 5}

The user prompt presents separate original and final node lists, including
IDs, types, semantic names, outline labels, page references, claims,
functions, and techniques. The system prompt asks for pairs or small groups
that state corresponding results, carry out corresponding proof steps, or
play the same argumentative role despite rewording or reorganization. It
permits split and merge proposals. Node types are treated as argumentative
roles rather than compatibility constraints: cross-type correspondences and
groups mixing statements with proof material are allowed when content has
genuinely been split, merged, absorbed, or externalized. When both documents
represent a result and its proof separately, however, the prompt prefers
specific statement-to-statement and proof-to-proof candidates over a coarse
theorem--proof package. It also discourages topical matches and coarse groups
already expressed by more specific proposals. Nodes may remain unmatched or
participate in distinct plausible candidates.

Deterministic statement anchors additionally supply associated proof pairs to
the model as weak hints worth examining. The prompt states that matching
parent statements do not establish that their proofs correspond. A proposal
involving proof material must instead identify at least one concrete shared
technique, intermediate claim, logical reduction, dependency, or substantive
proof step. Merely proving the same result is insufficient.

Each proposal contains nonempty \texttt{original\_ids} and
\texttt{final\_ids}, a \texttt{confidence} in $[0,1]$, and a one-sentence
\texttt{rationale} identifying what is shared. It also contains a structured
\texttt{shared\_proof\_features} list, whose entries classify proof-specific
evidence as a technique, intermediate claim, logical reduction, dependency,
or proof step. The model compares node descriptions; graph topology is not
included in this prompt. Proposals with unknown endpoints are discarded, as
are proof-related proposals that contain no proof-specific shared feature.
After proposals from all sources are combined, a coarse model-generated
statement-plus-proof package is suppressed when smaller direct statement and
proof candidates fully express the same endpoints.

\paragraph{Confidence calculation.}
For deterministic title anchors, let $s$ be the Jaccard similarity between the
two titles after tokenization and stop-word removal. Titles are eligible to
match when $s\geq 0.30$, and the resulting confidence is
$0.6+0.4s$. When each map contains exactly one result of a given outline type,
the type-uniqueness fallback proposes that pair even if the titles do not match;
it receives confidence $0.6$. A proof candidate derived from either kind of
statement anchor is only a weak parent-statement hint and receives $0.25$
times the statement candidate's confidence. For an imported passage pair,
let $q$ be the smaller of the two
passage-coverage fractions; its confidence is $0.4+0.4q$. The model supplies
the confidence for each model proposal directly as part of its structured
response. Its rubric treats $0.00$--$0.24$ as insufficient evidence,
$0.25$--$0.49$ as a weak correspondence in role or outer structure,
$0.50$--$0.74$ as a partial correspondence with a meaningful shared portion,
$0.75$--$0.89$ as a strong correspondence in central method or substantive
steps, and $0.90$--$1.00$ as near-equivalent argumentative content. Proofs
with materially different central methods must remain below $0.75$; higher
scores require concrete proof-specific commonality. When proposals with
identical endpoints are merged, the stored
confidence is their maximum and all proposal provenance is retained. Scores are
rounded to two decimal places. These values are proposer-specific, uncalibrated
heuristics rather than probabilities.

\paragraph{Correction prompts and provenance.}
Enrichment allows up to five targeted correction calls after the initial
request, for six total attempts, when mathematical text is corrupted or a
semantic definition name conflicts with its node type. Classification allows
one correction call for omitted,
duplicated, or extra pairs, or corrupted evidence, requesting a complete
replacement classification. Run logs retain the system prompt and its
version hash, the exact user prompts, model identity, raw responses, token
usage, and attempt information. Source ingestion, deterministic segmentation,
anchor scoring, passage grounding, deduplication, and visualization require
no model prompt. The Claude, Gemini, and GPT-5.6 Sol prompts used to construct
the illustrative proof pairs precede this pipeline; their roles are described
in Section~\ref{sec:examples}.

\section{The comparison interface}
\label{app:interface}

\begin{figure}[H]
  \centering
  \includegraphics[width=\linewidth]{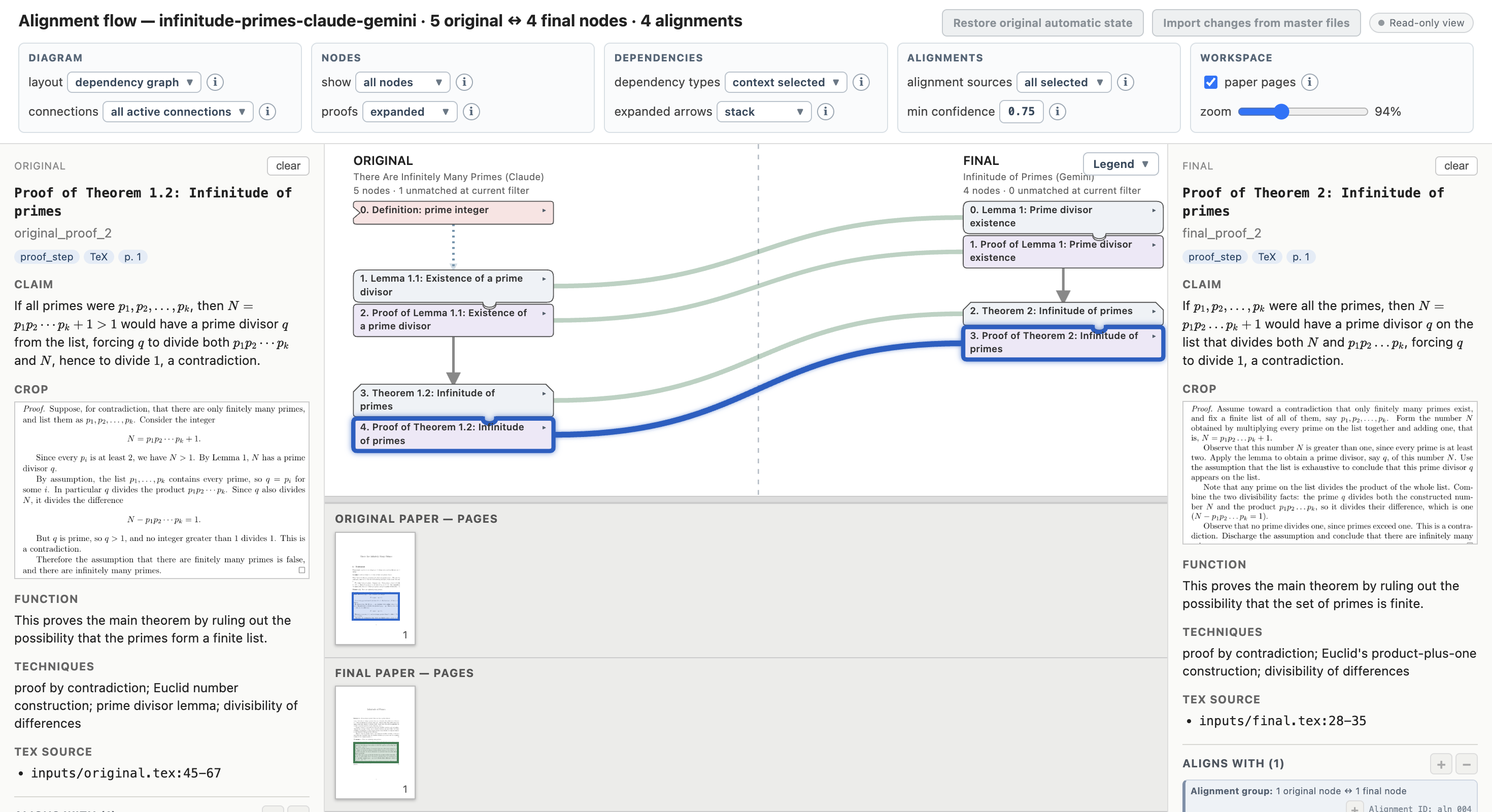}
  \caption{The \system{} comparison interface. The toolbar controls the
  central argument maps; the left and right panels describe selected
  original and final nodes and display their source evidence. This read-only
  view contains four automatic candidate alignments between five original and
  four final nodes. Proofs are expanded, the confidence threshold is $0.75$,
  and the paper-page view is enabled. Selecting the correspondence between the
  two theorem proofs highlights its endpoints in the maps, opens them in the
  side information panels, and marks their source regions in the original and
  final paper-page views. The original definition of a prime remains
  unmatched.}
  \label{fig:flow-interface}
\end{figure}

The browser-based \texttt{flow.html} interface connects the argument-map
overview to individual nodes and their source evidence. It is the read-only
output of step~6 in Figure~\ref{fig:pipeline-overview}; the same view is served
through the local \texttt{flow-edit} workflow when saved human curation is
needed. Figure~\ref{fig:flow-interface} shows the infinitude-of-primes
comparison with the corresponding theorem proofs selected, proof nodes
expanded, the minimum-confidence control set to $0.75$, and the paper-page
view enabled.

\subsection{Reading the central comparison}

The original document occupies the left side of the canvas and the final
document the right. Each card represents an argument node, with a heading
that combines its reading-order index and semantic label. Card shapes
distinguish node types. Expanded proofs appear as lavender cards attached
to their statements, and the small triangle in a card heading shows or
hides its claim summary. A thick outline marks a selected node. A pale red
statement card, such as the original definition in the figure, indicates
that the node has no candidate under the current filters; it does not
establish that the content is novel.

Thin directed edges within a document represent typed relations. Logical
edges point from a supporting node to a node that uses it, with transitive
reduction simplifying the drawing. Definition unfoldings use purple dashed
edges, while contextual relations use pale dotted edges. Forward
dependencies are marked in amber and dependency cycles in red.

Thicker curves crossing between documents represent candidate alignments.
Their width reflects confidence, and dashed curves indicate grouped
correspondences. Green, amber, and blue distinguish anchor, imported, and
model proposal sources. A candidate supported by several sources has one
display color; its list of sources is available in the inspection panel.
When collapsed proofs cause several links to share one visible curve, a
small count indicates the number of bundled candidates. Alignment coverage
and the unmatched counts above each map respond to the active filters.

\subsection{Toolbar controls}

Table~\ref{tab:interface-controls} summarizes the controls visible above the
maps. The confidence slider sets an inclusive minimum for displayed
candidates. Changing this minimum or enabling a proposal source changes
the displayed evidence; it does not accept or reject a correspondence.
The initial state uses the dependency-graph layout, all nodes with proofs
collapsed, all active connections with conceptual-context dependencies,
all alignment sources, $c\geq 0.75$, hidden paper pages, and 100\% zoom.
The \emph{expanded arrows} control applies only when proofs are expanded.
Its default \emph{Stack} mode treats a statement and its proof nodes as one
visual unit: an arrow runs from the bottom of the supporting stack to the top
of the dependent stack, and duplicate arrows between the same two stacks are
bundled. \emph{Detailed} mode instead connects the specific participating
nodes within those stacks.

\begin{table}[ht]
  \centering
  \small
  \begin{tabular}{@{}p{0.28\linewidth}p{0.66\linewidth}@{}}
    \toprule
    Control & Effect \\
    \midrule
    \raggedright Diagram: layout &
      Arrange nodes in source order or by logical dependence, placing
      prerequisites above the nodes that use them. \\[3pt]
    \raggedright Diagram: connections &
      Show all active connections, the original automatic result without human
      changes, or only saved manual additions, modifications, and rejections. \\[3pt]
    \raggedright Nodes: show &
      Show all nodes, only unmatched nodes, or focus the view by hiding nodes
      connected only through conceptual context, definition use, or both. \\[3pt]
    \raggedright Nodes: proofs &
      Collapse each proof into its claim card or expand claims and proofs into
      separate cards and alignment endpoints. \\[3pt]
    \raggedright Dependencies: types &
      Add definition-use and conceptual-context edges to the always-available
      result-use dependencies. \\[3pt]
    \raggedright Dependencies: expanded arrows &
      In expanded-proof view, draw bundled arrows between whole
      statement--proof stacks or detailed arrows between participating nodes. \\[3pt]
    \raggedright Alignments: sources &
      Display candidates supported by at least one enabled proposal source. \\[3pt]
    \raggedright Alignments: min confidence &
      Hide candidate links whose stored score is below the selected value. \\[3pt]
    \raggedright Workspace: paper pages &
      Show or hide the rendered-page strips below the comparison. \\[3pt]
    \raggedright Workspace: zoom &
      Resize the graph view; horizontal and vertical scrolling provide
      access to content outside the viewport. \\
    \bottomrule
  \end{tabular}
  \caption{Controls for exploring the comparison without changing the
  underlying argument maps or candidate set.}
  \label{tab:interface-controls}
\end{table}

\subsection{Inspecting nodes and source evidence}

Clicking a node opens its details in the corresponding side panel. Clicking
a cross-document curve opens the visible endpoint nodes in both panels, so
the selected alignment is reflected in the map and in both information panels.
Hovering over cards or curves highlights their connections. Selecting a
dependency or alignment row in an inspector emphasizes that connection in the
diagram. Each inspector shows the node's identifier, type, semantic source,
claim, argumentative function, and techniques. The \emph{TeX source} field
gives source-file and line locations. The \emph{Depends on} entries identify
supporting nodes, relation types, and textual evidence; contextual relations
and outgoing uses are listed separately when present.

The \emph{Candidates} list reports candidate identifiers, confidence,
alignment shape, counterpart node IDs, proposal sources, textual rationales,
and review status. This list includes all candidates involving the selected
node, including those hidden by the canvas filters. For a grouped
correspondence, the endpoint IDs identify the additional nodes to inspect.
Available source crops appear below the metadata. Each panel has a \emph{clear}
button and can be resized to provide more space for the mathematical text.
When selected nodes are aligned, the final-side inspector also provides a
collapsible word-level comparison of their claims and argumentative functions.

With \emph{paper pages} enabled, the lower strips display the original and
final PDFs. Selecting one node marks only that node's mapped location. Selecting
an alignment marks the original-paper region in blue and the final-paper region
in green, and scrolls both strips to the selected evidence. The page area can be
resized for closer inspection. These visual links are available when
source-to-page mappings have been produced.

\paragraph{Example inspection.}
In Figure~\ref{fig:flow-interface}, selecting the correspondence between the
two theorem proofs highlights the curve and its endpoint cards and opens the
two proofs in the side information panels. The panels show each claim, source
crop, TeX location, argumentative function, and techniques. Because the
paper-page view is enabled, the same selection also marks the corresponding
source region in blue on the original page and in green on the final page.
These linked views let the reader inspect the proposed correspondence against
both the structured node descriptions and the rendered papers. The unmatched
definition card shows a difference in explicit organization: only the original
map contains a separate definition node. Human reviewers may then open the pair through
\texttt{flow-edit} to curate dependencies and alignments as described below;
these revisions do not establish mathematical correctness.

\subsection{Human curation and revision history}

The ordinary \texttt{flow.html} output is read-only. The local
\texttt{flow-edit} service enables the side-panel controls that add, modify,
reject, or restore dependencies and alignments, and saves accepted decisions in
\texttt{flow\_curation.json}. A proposed edit is validated and presented as a
change summary before approval. Once approved, \system{} updates the active
comparison and deterministically recalculates affected maps and alignments
without another model call. Cancelling an edit restores the last accepted
state.

Manual alignments are initialized with confidence $1.0$, which reviewers may
edit. This manually assigned value is not a calibrated probability and is not
directly comparable to automatic proposer scores.

For a TeX-backed comparison, the curated state is stored in synchronized
editable master documents: one annotated TeX master for each input and one
pair-level alignment master. The TeX masters record node boundaries,
descriptions, and within-document dependencies without changing the papers'
text or rendering; the alignment master records the cross-document state.

For such TeX-backed comparisons, each accepted edit creates an immutable
pair-level revision. The editor can restore the complete pair---nodes,
dependencies, and alignments---to its first automatic revision while preserving
the immutable audit history. The stored history distinguishes automatic output
from human intervention but is not currently exposed as a user-facing version
browser.

%% file: references.bib
@article{thurston1994proof,
  author  = {Thurston, William P.},
  title   = {On Proof and Progress in Mathematics},
  journal = {Bulletin of the American Mathematical Society},
  volume  = {30},
  number  = {2},
  pages   = {161--177},
  year    = {1994},
  doi     = {10.1090/S0273-0979-1994-00502-6}
}

@misc{grobid,
  author       = {Lopez, Patrice and others},
  title        = {{GROBID}: A Machine Learning Software for Extracting
                  Information from Scholarly Documents},
  howpublished = {\url{https://github.com/grobidOrg/grobid}},
  year         = {2008},
  note         = {Software repository; accessed October 2026}
}

@inproceedings{blecher2024nougat,
  author    = {Blecher, Lukas and Cucurull, Guillem and Scialom, Thomas and
               Stojnic, Robert},
  title     = {Nougat: Neural Optical Understanding for Academic Documents},
  booktitle = {International Conference on Learning Representations},
  year      = {2024},
  url       = {https://openreview.net/forum?id=fUtxNAKpdV}
}

@inproceedings{mishra2024multimodal,
  author    = {Mishra, Shrey and Gauquier, Antoine and Senellart, Pierre},
  title     = {Modular Multimodal Machine Learning for Extraction of Theorems
               and Proofs in Long Scientific Documents},
  booktitle = {Proceedings of the 24th ACM/IEEE Joint Conference on Digital
               Libraries},
  year      = {2024},
  doi       = {10.1145/3677389.3702540}
}

@inproceedings{mansouri2022arqmath,
  author    = {Mansouri, Behrooz and Novotn{\'y}, V{\'i}t and Agarwal, Anurag
               and Oard, Douglas W. and Zanibbi, Richard},
  title     = {Overview of {ARQMath-3} (2022): Third {CLEF} Lab on Answer
               Retrieval for Questions on Math},
  booktitle = {Working Notes of CLEF 2022},
  series    = {CEUR Workshop Proceedings},
  volume    = {3180},
  year      = {2022},
  url       = {https://ceur-ws.org/Vol-3180/paper-35.pdf}
}

@inproceedings{li2023bert,
  author    = {Li, Weixian Waylon and Ziser, Yftah and Coavoux, Maximin and
               Cohen, Shay B.},
  title     = {{BERT} Is Not The Count: Learning to Match Mathematical
               Statements with Proofs},
  booktitle = {Proceedings of the 17th Conference of the European Chapter of
               the Association for Computational Linguistics},
  pages     = {3581--3593},
  year      = {2023},
  doi       = {10.18653/v1/2023.eacl-main.260}
}

@misc{kurgan2026theoremgraph,
  author        = {Kurgan, Simon and Wang, Evan and Leonen, Eric and Szeto,
                   Sophie and Alexander, Luke and Remizov, Artemii and Alper,
                   Jarod and Inchiostro, Giovanni and Ilin, Vasily},
  title         = {{TheoremGraph}: Bridging Formal and Informal Mathematics},
  year          = {2026},
  eprint        = {2606.25363},
  archiveprefix = {arXiv},
  doi           = {10.48550/arXiv.2606.25363},
  url           = {https://arxiv.org/abs/2606.25363}
}

@inproceedings{lauscher2018sciarg,
  author    = {Lauscher, Anne and Glava{\v{s}}, Goran and Ponzetto, Simone Paolo
               and Eckert, Kai},
  title     = {Investigating the Role of Argumentation in the Rhetorical
               Analysis of Scientific Publications with Neural Multi-Task
               Learning Models},
  booktitle = {Proceedings of the 2018 Conference on Empirical Methods in
               Natural Language Processing},
  pages     = {3326--3338},
  year      = {2018},
  doi       = {10.18653/v1/D18-1370}
}

@article{ruosch2025interdocument,
  author  = {Ruosch, Florian and Sarasua, Cristina and Bernstein, Abraham},
  title   = {Mining Inter-Document Argument Structures in Scientific Papers
             for an Argument Web},
  journal = {Transactions on Graph Data and Knowledge},
  volume  = {3},
  number  = {3},
  pages   = {4:1--4:33},
  year    = {2025},
  doi     = {10.4230/TGDK.3.3.4}
}

@misc{petrov2025proof,
  author        = {Petrov, Ivo and Dekoninck, Jasper and Baltadzhiev, Lyuben and
                   Drencheva, Maria and Minchev, Kristian and Balunovi{\'c},
                   Mislav and Jovanovi{\'c}, Nikola and Vechev, Martin},
  title         = {Proof or Bluff? Evaluating {LLM}s on 2025 {USA} Math
                   Olympiad},
  year          = {2025},
  eprint        = {2503.21934},
  archiveprefix = {arXiv},
  url           = {https://arxiv.org/abs/2503.21934}
}

@inproceedings{godlin2009regression,
  author    = {Godlin, Benny and Strichman, Ofer},
  title     = {Regression Verification},
  booktitle = {Proceedings of the 46th Annual Design Automation Conference (DAC)},
  pages     = {466--471},
  year      = {2009},
  doi       = {10.1145/1629911.1630034}
}

@inproceedings{ringer2021proofrepair,
  author    = {Ringer, Talia and Porter, RanDair and Yazdani, Nathaniel and Leo, John
               and Grossman, Dan},
  title     = {Proof Repair across Type Equivalences},
  booktitle = {Proceedings of the 42nd ACM SIGPLAN International Conference on
               Programming Language Design and Implementation (PLDI)},
  year      = {2021},
  doi       = {10.1145/3453483.3454033}
}

@inproceedings{first2023baldur,
  author    = {First, Emily and Rabe, Markus N. and Ringer, Talia and Brun, Yuriy},
  title     = {Baldur: Whole-Proof Generation and Repair with Large Language Models},
  booktitle = {Proceedings of the 31st ACM Joint European Software Engineering
               Conference and Symposium on the Foundations of Software Engineering
               (ESEC/FSE)},
  pages     = {1229--1241},
  year      = {2023},
  doi       = {10.1145/3611643.3616243}
}

@misc{leanrefactorarena2026,
  author       = {{VeriCodeGen Organizers}},
  title        = {{Lean Refactor Arena}},
  howpublished = {Competition track of the {NeurIPS} 2026 Workshop on {AI} for
                  Verifiable Coding ({VeriCodeGen}),
                  \url{https://leanrefactor.github.io/}},
  year         = {2026},
  note         = {Accessed October 2026}
}
